\documentclass[runningheads]{llncs}
\usepackage{graphicx}
\usepackage{amsmath,amssymb} 
\usepackage{color}
\usepackage[width=122mm,left=12mm,paperwidth=146mm,height=193mm,top=12mm,paperheight=217mm]{geometry}

\usepackage[utf8]{inputenc} 
\usepackage{hyperref}       
\usepackage{url}            
\usepackage{booktabs}       
\usepackage{amsfonts}       
\usepackage{nicefrac}       
\usepackage{microtype}      
\usepackage{wrapfig}

\usepackage{subfig}

\usepackage{graphicx}
\graphicspath{{Images/}}

\newcommand{\etal}{\emph{et al.\ }}
\newcommand{\eg}{e.g.\ }
\newcommand{\ie}{i.e.\ }

\begin{document}

\title{Attaining human-level performance with atlas location autocontext for anatomical landmark detection in 3D CT data} 

\titlerunning{Atlas location autocontext for anatomical landmark detection in 3D CT data}

\author{
Alison Q. O'Neil\inst{1}\and
Antanas Kascenas\inst{1}\and
Joseph Henry\inst{1}\and
Daniel Wyeth\inst{1}\and
Matthew Shepherd\inst{1}\and
Erin Beveridge\inst{1}\and
Lauren Clunie\inst{1}\and
Carrie Sansom\inst{1}\and
Evelina \v Seduikyt\.e\inst{1}
Keith Muir\inst{2}\and
Ian Poole\inst{1}
}


\authorrunning{Alison Q. O'Neil \etal}



\institute{
Canon Medical Research Europe, Edinburgh EH6 5NP \and
Queen Elizabeth University Hospital, University of Glasgow, Glasgow G51 4TF}

\maketitle

\begin{abstract}
We present an efficient neural network method for locating anatomical landmarks in 3D medical CT scans, using atlas location autocontext in order to learn long-range spatial context. Location predictions are made by regression to Gaussian heatmaps, one heatmap per landmark. This system allows patchwise application of a shallow network, thus enabling multiple volumetric heatmaps to be predicted concurrently without prohibitive GPU memory requirements. Further, the system allows inter-landmark spatial relationships to be exploited using a simple overdetermined affine mapping that is robust to detection failures and occlusion or partial views. Evaluation is performed for 22 landmarks defined on a range of structures in head CT scans. Models are trained and validated on 201 scans. Over the final test set of 20 scans which was independently annotated by 2 human annotators, the neural network reaches an accuracy which matches the annotator variability, with similar human and machine patterns of variability across landmark classes.
\end{abstract}

\section{Introduction}

By ``anatomical landmark detection'', we refer to the task of detecting and localising points in the human body which can be uniquely defined in terms of the anatomical landscape, for instance \emph{superior aspect of right eye globe} or \emph{base of pituitary gland}. Landmark identification is an important enabling technology, providing semantic information that can be used to initialise or aid other medical image analysis algorithms, such as volume registration \cite{Rohr2001,Hellier2003,Polzin2014,Han2015}, organ segmentation \cite{Chen2014,Ibragimov2014,Ibragimov2012,Lay2013}, vessel tracking \cite{ONeil2014}, computer aided detection of pathology \cite{Zhang2016,Lisowska2017}, treatment planning \cite{Leavens2008}, and therapy assessment \cite{Dong2015}.

Taking a machine learning approach to automated detection enables the heterogeneity of appearance of each landmark to be conveniently represented. Fully convolutional neural networks (FCNs) are particularly well suited to this task, since whole volumes may be efficiently parsed to detect and localise multiple landmark points concurrently using a learned, shared feature representation.

For the purposes of prediction, the concept of a landmark may be modelled in different ways. An intuitive method would be to regress the positions of the landmarks. This can be done by training the network to make voxelwise predictions of the Euclidean offsets of all landmarks, as in \cite{Gao2014,Riegler2014}, then using a scheme such as Hough regression to combine the votes. Offset regression carries a heavy learning burden, since the network must learn to recognise every voxel in a scan, or at least sufficient voxels to enable voting by agreement, and make precise, subtly differing long-range spatial predictions, mapping appearance features to distance measures in the process (\ie ``Where am I relative to each landmark of interest?''). An alternative, more lightweight method is the heatmap regression technique of Payer \etal \cite{Payer2016} in which the network is trained to predict the presence of Gaussian heat spots centred at the landmark locations; this is mathematically equivalent to learning a nonlinear measure of the Euclidean landmark offset \emph{magnitude} and is a simpler learning task much more akin to straightforward appearance matching (\ie ``How much do I look like each landmark of interest?'').

An important element of the landmark detection problem is how to incorporate long-range spatial context, since points in different parts of the body may have similar appearance and thus be confounded. In \cite{Payer2016}, the initial appearance-based CNN is followed by a ``spatial configuration unit'' in which each landmark predicts the location of every other landmark by learning the relative Euclidean offset. This is a reasonable approach for the featured problem of hand X-Ray images, however it would not scale well to body parts and scan protocols in which the orientation, scale and acquisition region are variable. Other methods of capturing global context include U-Net \cite{Ronneberger2015} (or the similar V-Net \cite{Milletari2016}), dual-pathway approaches \cite{Kamnitsas2017,Fan2015}, dual networks \cite{Lu2016}, iterative cascaded networks \cite{Toshev2014}, and the reinforcement learning method of Ghesu \etal \cite{Ghesu2017}. These methods describe various mechanisms for learning both local and long-range information. The U-Net and dual-pathway approaches are methods of combining information at different resolutions in a single end-to-end trained network, whilst the dual network approach delegates the learning of global and local context to different networks. The approach of Toshev and Szegedy \cite{Toshev2014} is similar to Tu and Bai's idea of autocontext \cite{Tu2010}, in which the network predictions are iteratively fed to subsequent networks along with the image data such that context is gradually learnt, or gathered, into the network model. Finally, Ghesu \etal's reinforcement learning approach involves the navigation of multiple agents through a scan volume (from different starting points) until they converge on the landmark position. Thus, agents explicitly train to be spatially aware. A drawback of this approach is its lack of scalability and the potential redundancy since each landmark requires a separate model to be trained.

This paper builds on the work of O'Neil \etal \cite{ONeil2015,ONeil2016} in which Tu and Bai's idea of autocontext \cite{Tu2010} (iteratively feeding the probabilistic output of a model to a subsequent model) was modified to \emph{atlas location} autocontext (iteratively feeding the coordinate in atlas space, according to the output of a model, to a subsequent model). In these previous works, a decision forest was used. In this paper we show that the decision forest can be replaced by a shallow fully convolutional neural network, which outperforms the decision forest method, and attains human-level performance. Since the model is shallow, this system is memory and time efficient. Memory efficiency is particularly important when taking a unified approach for problems with large 3D inputs and many output classes (many landmarks), requiring many kernels throughout the network, including in the final layers.

\section{Method}

\subsection{Landmark Detection System}

\subsubsection{Atlas Location Autocontext}

\begin{figure}[ht]
  \centering
  \fbox{\includegraphics[width=0.85\textwidth]{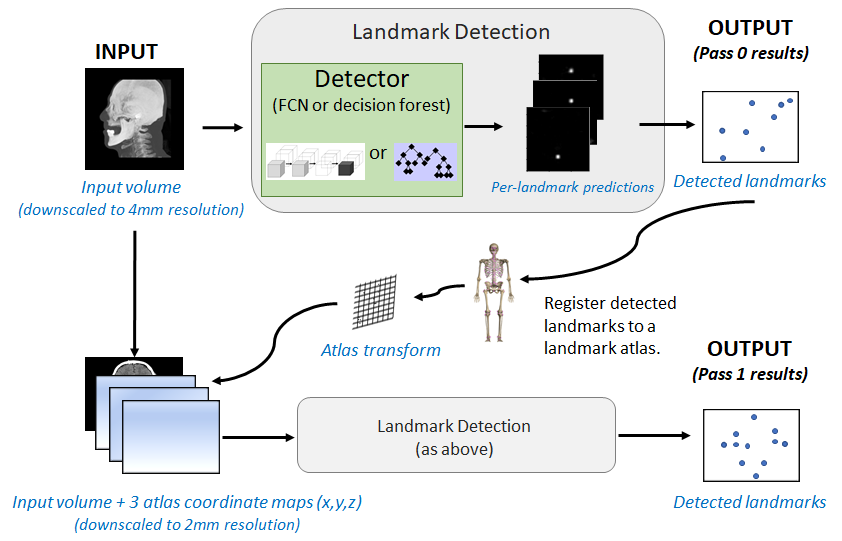}}\\
  \caption{Landmark detection system using atlas location autocontext}
	\label{fig:AtlasFeedback}
\end{figure}

The landmark detection system consists of a cascade of two models, with the output of the first providing spatial information to the second in the form of estimated $x$, $y$ and $z$ atlas space coordinates. The second detector can then be trained not only on image intensity features but also on approximate spatial features; this transmission of learned contextual information is what we term atlas location autocontext. The two models have identical architectures, except that the first has 1 input (image) and the second has 4 inputs (image +  atlas space coordinates). We choose to train the first model with data at lower resolution (4mm per voxel) and the second at higher resolution (2mm per voxel) in order to emphasise learning of spatial context in pass 0, and learning of local appearance in pass 1. See Figure \ref{fig:AtlasFeedback} for illustration.

Coordinates are determined in this paper by affine alignment of the first model's predicted landmark locations to a landmark atlas, using \emph{iterative weighted least squares fitting}. The least squares fit is that which minimises the sum of the squared distances between the atlas landmarks and the mapped detected landmarks. Since the detected landmarks will sometimes be erroneous or innaccurate --- hence the need for a second model! --- we weight distances by their detection \emph{certainty} values (see Section \ref{sec:Inference}) to prioritise fitting of the more confident detections, and then we do iterative refinement. Iterative refinement involves removing landmarks one at a time \ie dropping the landmark with the largest mapping error, and subsequently recomputing the mapping, until all remaining (mapped) detected landmarks are within a distance $d_{Atlas}$ of the corresponding atlas landmarks. In this way a subset of landmark predictions is discovered with a plausible spatial configuration. The value for $d_{Atlas}$ was chosen by parameter sweep to minimise the average mapping error across the training scan results.

\subsubsection{Direct Atlas Correction}

For additional robustness, we directly leverage the affine atlas mapping to correct outliers, by mapping the atlas landmarks back to the volume and adjusting each landmark's predicted location to be the voxel with maximum certainty within a distance $d_{Volume}$ of its mapped atlas counterpart. In other words, we generate spherical regions of interest (ROIs) with radius $d_{Volume}$, within which the detections must lie. This allows correction of conspicuous outliers and on its own could perhaps be considered a cheap form of ``autocontext''. For this step we select a generous threshold of $d_{Volume}$ = 28mm.

\subsection{Proposed FCN}

\subsubsection{Model Architecture}

\begin{figure}[ht]
  \centering
  \fbox{\includegraphics[width=0.97\textwidth]{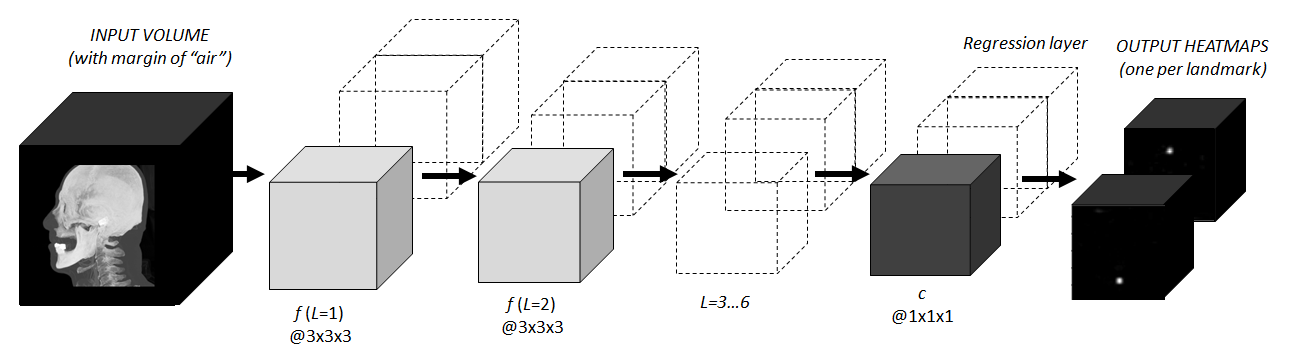}}\\
  \caption{Our proposed fully convolution network. The number of filters $f(L) = a \times 2^L$, for $a=12$ and layers $L = 0, 1, 2, ... 5$.}
	\label{fig:FcnArchitecture}
\end{figure}

The model has a straightforward architecture (see Figure \ref{fig:FcnArchitecture}), with 6 layers of $3 \times 3 \times 3$ kernels, where there are 12 kernels in the first layer and the number of kernels doubles in every subsequent layer. The model has 2,661,166 parameters in total. All convolutions are performed using ``valid mode'' (\ie the input shrinks at each convolution) and use ReLU activation functions except for the final regression layer which has a linear activation.

\vspace{-20pt}

\subsubsection{Data pre-processing}

Voxel intensities were normalised by first rescaling the HU intensities by 3 x $10^{-3}$ since this puts the soft tissue values in the typical $[-1, 1]$ range, and then truncating values to fit the range $[-3, 3]$ (\ie [-1000HU, 1000HU]). In order to detect landmarks at the edge of the scan, each scan was dilated by the size of the margin required by the network, using pixels with a value equivalent to air (as opposed to zero padding). During training, the data was augmented by left-right reflection of the volume with corresponding switching of the left and right side landmarks. To introduce robustness to acquisition region, scans were randomly cropped with a margin of up to 50mm. In practice this was done by uniformly sampled translations in the range +/-50mm in $x$, $y$ and $z$.

Patches were used during both training and application. At training time, this was done in order to control for data imbalance and also (pragmatically) to allow samples from many volumes within each batch without large memory requirements. Patches of $15 \times 15 \times 15$ (\ie predictions made for the central $3 \times 3 \times 3$ voxels plus the 6-voxel margin required by the model) were extracted at the landmark positions as well as randomly from the remainder of the volume at a ratio of 1:5. At application time, patches of $30 \times 30 \times 30$ (\ie $42 \times 42 \times 42$ including margin) were tiled to make piecewise predictions covering the whole volume.

\subsubsection{Inference}
\label{sec:Inference}

To make the predictions, we use a modified version of the heatmap regression proposed by Pfister \etal \cite{Pfister2015} and applied previously to landmark detection in medical scans by Payer \etal \cite{Payer2016}. In this scheme each landmark has a separate volumetric output containing a Gaussian heat spot centred at the landmark position. More formally, the temperature $t_i$ of the $i$th heatmap that we regress against is determined according to distance of the voxel $v$ from the landmark position $p_i$ for landmark $i$, a standard deviation $\sigma$ (chosen to be 1 voxel), and a constant $k$ denoting the Gaussian height:

\begin{equation}
t_i = ke^{-{\frac{(v - p_i)^2}{2\sigma^2}}}
\end{equation}
\vspace{-20pt}
\\

We used mean squared error as the loss function and found empirically that the imbalance between background and proximal landmark voxels meant large heights were required for the Gaussian in order to enable training to start (\ie $k$ = 1 x $10^3$ at 4mm resolution, and $k$ = 1 x $10^6$ at 2mm resolution). This mechanism was chosen for convenience; note that we could have equivalently initialised the network with small kernel weight initialisations, or experimented with weighting of the landmark voxels in the loss function. At prediction time, the predicted position for each landmark is chosen to be simply the output voxel with maximum value $t$. We divide $t$ by $k$ such that it lies in the range [0, 1], and we term this the landmark \emph{certainty}.

\subsubsection{Training procedure}

For each model, kernel weights were initialised using normalised He initialisation \cite{He2015}. Optimisation of the network was performed using backpropagation with Adam \cite{Kingma2014}, with learning rate=0.001, $\beta_1$ = 0.9 and $\beta_2$ = 0.999. Batches of 32 patches were used, and training was run for 50 epochs (first pass) and 200 epochs (second pass). The weights were retained from the epoch which achieved lowest error on the validation data.

\subsection{Benchmarking: Decision forest algorithm}

As our baseline for comparison, we follow the decision forest approach of Dabbah \etal \cite{Dabbah2014} for which we have a mature C++ implementation. Until the recent popular adoption of CNN solutions, decision forests and their variants were the gold standard for the task of anatomical landmark detection \cite{Jimenez-Del-Toro2016,Oktay2017,Oktay2017,Dai2015,Gao2016,Han2015,Zhang2016,Wang2015}. In brief, a decision forest is trained to perform voxelwise classification across $n + 1$ classes, where there are $n$ landmarks and 1 background class. Voxels $v_i$ within 1.5 voxels of the each landmark location $p_i$ are considered to be landmark samples, and are assigned a weight $w$ during training according to Gaussian distribution \ie:

\begin{equation}
w = ke^{-{\frac{(v_i - p_i)^2}{2\sigma^2}}}
\end{equation}
\vspace{5pt}\\

\noindent{}where $k = 1$ and $\sigma = 0.75$ voxels. Voxels outside of these spheres are considered to be background samples. The features for each voxel are the Hounsfield Unit (HU) values of the voxels in the local $100^3$mm neighbourhood (note that scan intensities are not normalised as was done for the FCN), with each tree being given a random sample of 2500 features (random subspace sampling \cite{Ho1998}). We trained 100 trees, sampling from $n$ = 40 randomly chosen training scans per tree. We further tried using HOG \cite{Dalal2005} features alongside the intensity features, since these had previously been shown to give improvement over using intensity features alone \cite{ONeil2016} (note that we used signed rather than unsigned orientations, with no magnitude weighting, as in \cite{ONeil2016}). In this case we randomly selected 1250 intensity features and 1250 HOG features per tree. Histograms were computed over randomly generated box regions of up to 48mm in each dimension.

At application time, the novel volume is scanned, and for each landmark, the voxel is selected which has the highest probability of belonging to that landmark class.

\section{Experiments}



\subsection{Data}

We demonstrate our method on CT head scan volumes. The data is split into 170 scans for training, 31 scans for validation, and a final (tested once) test set of 20 scans. The data was acquired from a range of scanners (Canon, Siemens, G.E., Philips), scan protocols (both with and without injected arterial contrast), and with a range of resolutions and slice thicknesses. There are approximately equal splits between male and female subjects. Many contain pathology, inclusive of haemorrhage, tumours and age-related change.

\begin{figure}[ht]
  \centering
  \fbox{\includegraphics[width=0.97\textwidth]{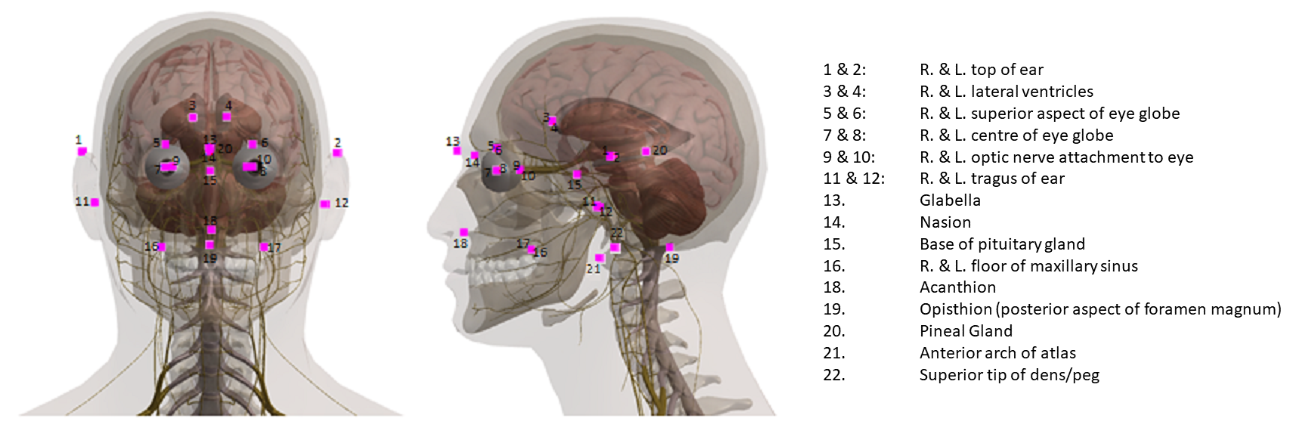}}\\
  \caption{Schematic of head landmarks}
	\label{fig:LandmarkSetHead}
\end{figure}

A set of 22 landmarks were defined in the head (see Figure \ref{fig:LandmarkSetHead}). Scan protocols were designed by an in-house clinical analyst (E.B.) with postgraduate-level expertise in anatomy. Three additional observers with education in biological sciences were trained up to perform the annotation. The test set was annotated by two observers, one of whom (observer A, L.C.) has also annotated a large number of the training scans, the second of whom (observer B, E.S.) was independent of the training data. In many scans, only a subset of the landmarks is visible. This may be either because the landmark lies outside of the scan acquisition region, or because the landmark is obscured for some reason, for instance low resolution data, the presence of pathology or the absence of contrast. In the latter case, it is marked as ``uncertain'' in the ground truth; the 6 landmarks which were marked as uncertain by at least one observer are not included in our metrics.

\subsection{Implementation}

The FCN was implemented in Python using the Keras library \cite{Chollet2015}, built on top of the Tensorflow library \cite{Abadi2016}. Parameter exploration was performed on p2.xlarge instances on AWS; these instances have one NVIDIA K80 GPU (2496 cores, 12 GB VRAM). On a computer with an NVIDIA GTX Titan X, run times are of the order of 1 second for the first pass and 2 seconds for the second (excluding data loading and downscaling, and model loading from disk). In the second pass, we reduce the run time by evaluating only parts of the volume containing the atlas-mapped spherical landmark ROIs identified in the first pass.

The decision forest is implemented in C++. Experiments were run on a computer with two Intel Xeon E5645 (2.4GHz) processors. Run times are of the order of 1 second for the first pass and 0.5 seconds for the second pass (excluding data loading and downscaling, and classifier loading from disk). There are a number of optimisations, as described in \cite{Dabbah2014}, for instance evaluating only as many trees as required until confident about the prediction, and performing coarse-to-fine scanning of the volume within a pass (\ie evaluate every second voxel in the volume before evaluating all voxels in the neighbourhood of the maximum-probability landmark positions).

\subsection{Results}

\begin{figure}[!ht]
\centering
\subfloat{\includegraphics[height=2.5cm]{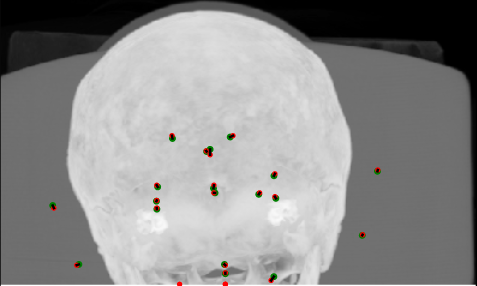}}\hspace{5pt}
\subfloat{\includegraphics[height=2.5cm]{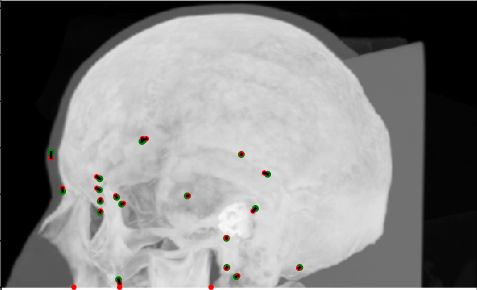}}\hspace{5pt}
\subfloat{\includegraphics[height=2.5cm]{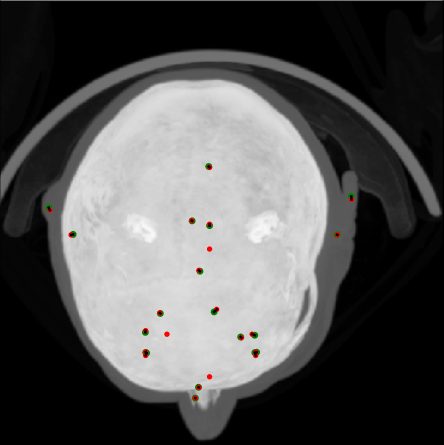}}\\
\subfloat{\includegraphics[height=2.5cm]{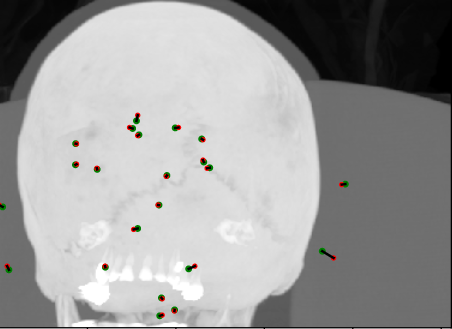}}\hspace{5pt}
\subfloat{\includegraphics[height=2.5cm]{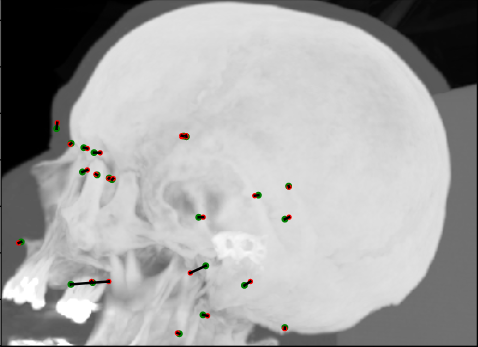}}\hspace{5pt}
\subfloat{\includegraphics[height=2.5cm]{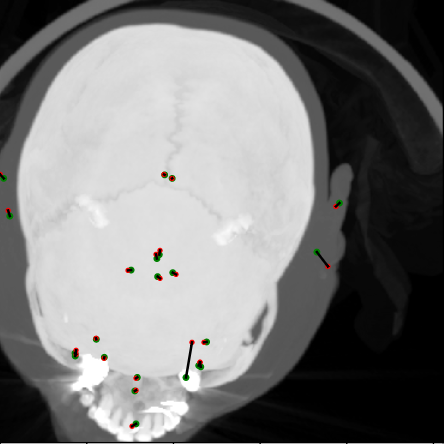}}\\
\vspace{0.2cm}
\caption{Coronal, sagittal and axial maximum intensity projections (MIPs) of results for a good case (top) and a poor case (bottom). Green dots = ground truth (observer A), red dots = detected (proposed FCN), and black lines connect corresponding pairs.}
\label{fig:HeadResultImages}
\end{figure}

Summary metrics are shown in Table \ref{Tab:Results} for landmark localisation errors (or disagreements), and some visual results are shown for the FCN in Figures \ref{fig:HeadResultImages} and \ref{fig:HeadHeatmapResultImages}. The summary metrics show that the FCN outperforms the decision forest methods. The anomalous metric is the mean \emph{max error}, in other words, the mean size of the ``worst detected landmark in a scan''. This does not appear to improve in the second pass --- if anything, the worst error worsens--- and the Pass 0 decision forest with HOG features is the best performer. We propose that this occurs because landmarks with atypical appearance (\eg see the calcified pineal gland example in Figure \ref{fig:HeadHeatmapResultImages}) are best located by use of spatial context rather than local appearance, hence the efficacy of low resolution HOG features which are aggregated over regions and thus are relatively insensitive to precise changes.

\begin{table}[t]
  \caption{Landmark localisation disagreement. The mean, median and max error metrics are computed over for each scan separately and then the mean value is taken across the 20 scans and provided below. Additionally we show the percentage (\%) of landmarks with an error greater than 4mm, computed over all 417 landmarks in the 20 scans. \emph{DF} = Decision Forest with intensity features, \emph{DF (+HOG)}= Decision Forest with intensity + HOG features, \emph{FCN} = proposed fully convolutional network.}
  \label{Tab:Results}
  \centering
	\vspace{0.2cm}
  \begin{tabular}{lcccccccc}
    \toprule
    \textbf{Method} & \multicolumn{8}{c}{\textbf{Reference}}\\
		\cmidrule{1-9}
		& \multicolumn{4}{c}{\textbf{Observer A}} & \multicolumn{4}{c}{\textbf{Observer B}} \\
															& Mean 	& Median & Max & \% & Mean & Median & Max & \% \\
    \midrule
		Observer A								& -  		& -  	 & -		&		-		& 2.20 	& 1.49  & 9.27	&	11.0 \\
		Observer B								& 2.20 	& 1.48 & 9.27	&	11.0	& 	-		& -  		& -			&		-	 \\
		\midrule
		\textbf{Pass 0 (4mm)} 		& & & & & & & & \\
		\midrule
		DF					 							& 4.47 & 4.03 & 11.54 & 50.4 & 4.58 & 4.14 & 11.28 & 49.2 \\
		DF (+HOG)									& 4.25 & 3.91 & 10.07 & 47.7 & 4.36 & 3.92 & 9.86 & 46.5 \\
		FCN												& 3.38 & 2.65 & 12.20 & 21.6 & 3.52 & 2.71 & 12.15 & 23.7 \\
		FCN	+ Atlas Correction		& 3.03 & 2.53 &	10.45	& 16.1 	& 3.31 & 2.62 &	10.89	&	21.6 	\\
		\midrule
		\textbf{Pass 1 (2mm)} 		& & & & & & & & \\
		\midrule
		DF												& 3.59 & 2.85 & 13.73 & 26.6 & 3.83 & 3.02 & 13.59 & 27.6 \\
		DF (+HOG)									& 3.30 & 2.88 & \textbf{9.77} & 24.9 & 3.47 & 2.84 & \textbf{9.69} & 26.9 \\
		FCN												& 2.93 & 1.50 & 19.98 & 12.2 & 3.42 & 1.84 & 20.93	&	17.5 \\
		FCN	+ Atlas Correction		& \textbf{2.29} & \textbf{1.49} & 11.41 & 10.8 & \textbf{2.77} & \textbf{1.78} & 12.26	&	\textbf{16.1} \\
		\midrule
		\multicolumn{9}{l}{\textbf{Alternative: Pass 0 (2mm) + Atlas Correction}}\\
		\midrule
		FCN 											& 2.55 & 1.55 &	15.08	& \textbf{10.3} & 3.10 & 1.92 &	15.38	&	17.5	\\
		\bottomrule
  \end{tabular}
	\vspace{0.2cm}
\end{table}

The significance of the improvements of the FCN over the decision forest were verified using a one-tailed paired Student's $t$-test  for the 417 landmark examples, using each observer in turn as the reference, and significance was found to hold for a $p$-value $<$ 0.01 for all comparisons. However, the results of the FCN model using atlas location autocontext + direct atlas correction were \emph{not} significantly different to those using only the direct atlas correction at 2mm resolution. It might be that significance could be shown with a larger population of datasets, or for landmarks which vary their relative position more dramatically relative to other structures (\eg on vessels); in this case the learning could learn the spatial distribution and adapt its localisation to the observed anatomical landscape where explicitly imposed affine constraints could not. What the autocontext system \emph{does} offer is a run time speed-up, since high-resolution processing can be performed selectively, as opposed to over the whole volume (mean run time of approximately 3 seconds \ie 1 + 2, as opposed to 5 seconds for the single-pass system) --- however in this case it seems that the atlas channels of the second model have not been conclusively proven to add a benefit. 

Regarding human vs. machine performance, the FCN achieves similar mean and median agreement with observer A as the agreement between observers A and B. However, the FCN is less well in agreement with observer B than observer A. There may be two reasons for this. Firstly, the algorithm was trained on annotations from observer A amongst others, so may have learned to mimic the annotation style of observer A. Secondly, since observer A was part of our team for training data annotation, all of her annotations were subject to our selective review process (a percentage of our ground truth observations are reviewed by E.B. for quality control). Therefore mistakes or inconsistencies are more likely to be have been picked up and corrected for observer A (\ie observer A's ground truth will have some of the characteristics of consensus ground truth).

\begin{figure}[!ht]
\centering
\includegraphics[height=5.5cm]{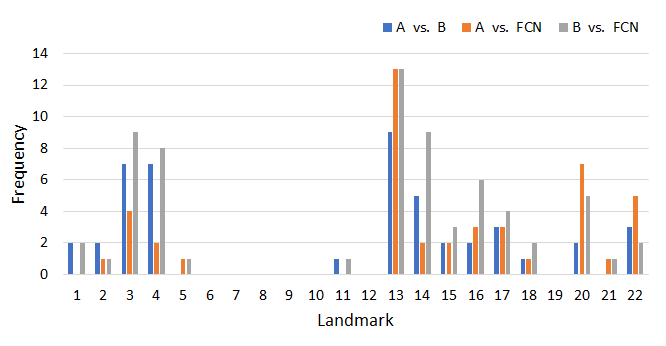}
\caption{Distribution of errors ($>$ 4mm) for all landmarks 1-22 (numbers correspond to those in Figure \ref{fig:LandmarkSetHead}). The pattern of errors is similar for human vs. human as for human vs. machine.}
\label{fig:PerLandmarkErrors}
\end{figure}

We further take those errors which are greater than 4mm, and show the breakdown between observers and between landmarks in Figure \ref{fig:PerLandmarkErrors}. There is a similar pattern to the human vs. human and the human vs. machine disagreement, with most discrepancies arising on surface landmarks (notably 13. = glabella, 3. \& 4. = L \& R frontal horns of the lateral ventricles). Landmarks on surfaces may be less well defined and inspection of the underlying predictions (see Figure \ref{fig:HeadHeatmapResultImages}) supports this. Other mistakes by the algorithm are due to landmark appearances less frequently (or never) seen in the training data, such as the calcified appearance of the pineal gland (20.) example in Figure \ref{fig:HeadHeatmapResultImages}.

\newcommand\x{0.135}

\begin{figure}[!ht]
\centering
\subfloat{\includegraphics[width=\x\textwidth]{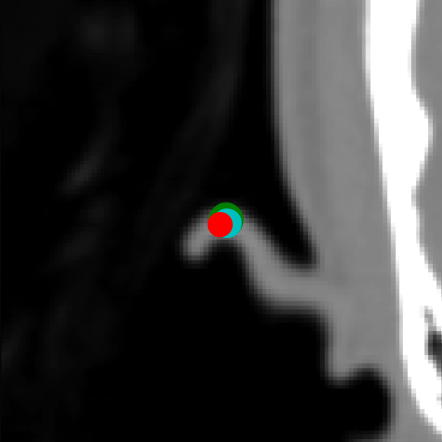}}\hspace{5pt}
\subfloat{\includegraphics[width=\x\textwidth]{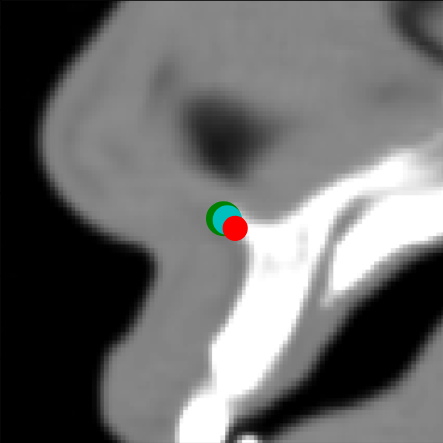}}\hspace{5pt}
\subfloat{\includegraphics[width=\x\textwidth]{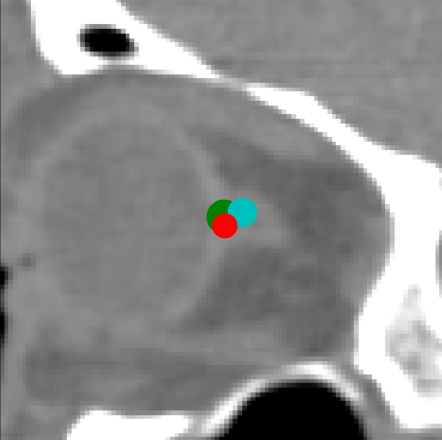}}\hspace{5pt}
\subfloat{\includegraphics[width=\x\textwidth]{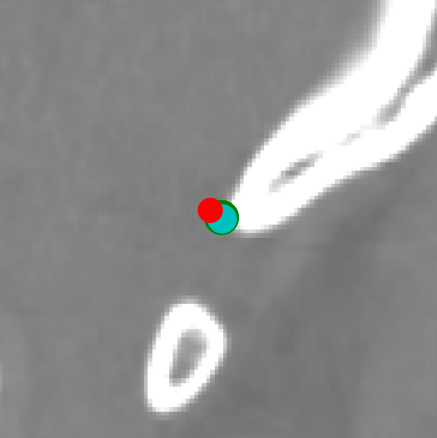}}\hspace{5pt}
\subfloat{\includegraphics[width=\x\textwidth]{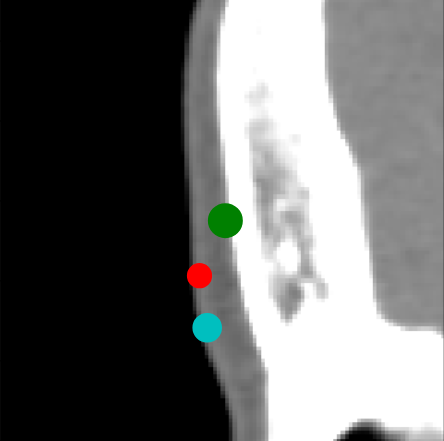}}\hspace{5pt}
\subfloat{\includegraphics[width=\x\textwidth]{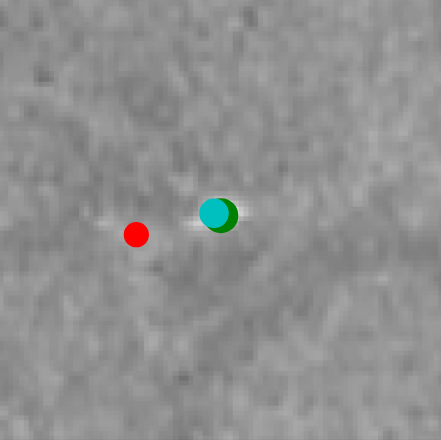}}\\
\subfloat{\includegraphics[width=\x\textwidth]{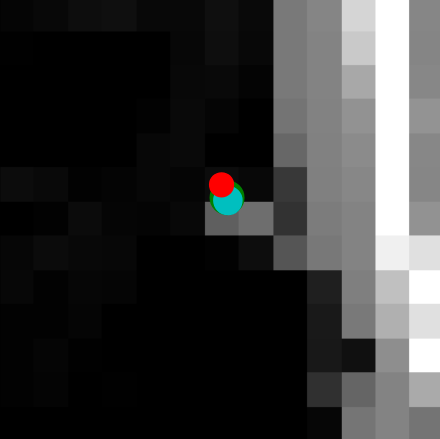}}\hspace{5pt}
\subfloat{\includegraphics[width=\x\textwidth]{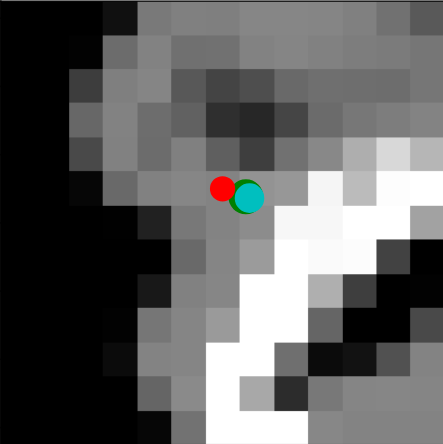}}\hspace{5pt}
\subfloat{\includegraphics[width=\x\textwidth]{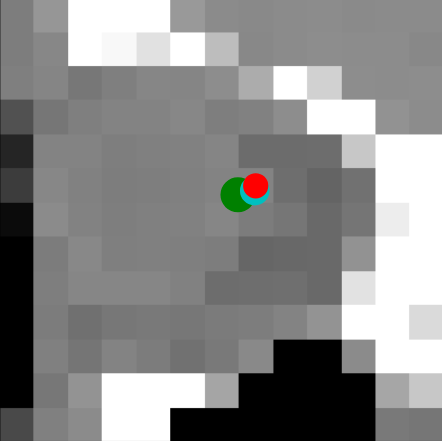}}\hspace{5pt}
\subfloat{\includegraphics[width=\x\textwidth]{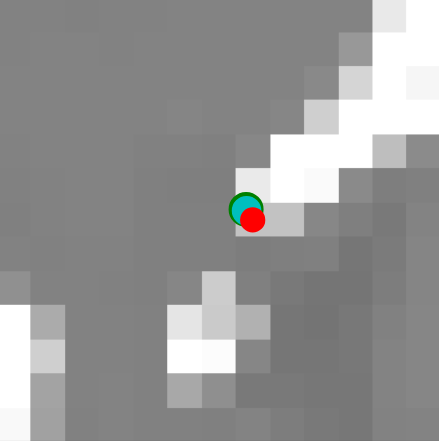}}\hspace{5pt}
\subfloat{\includegraphics[width=\x\textwidth]{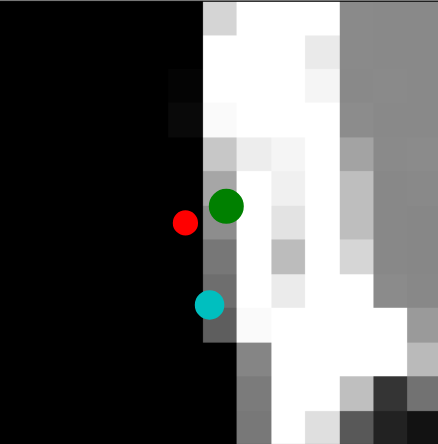}}\hspace{5pt}
\subfloat{\includegraphics[width=\x\textwidth]{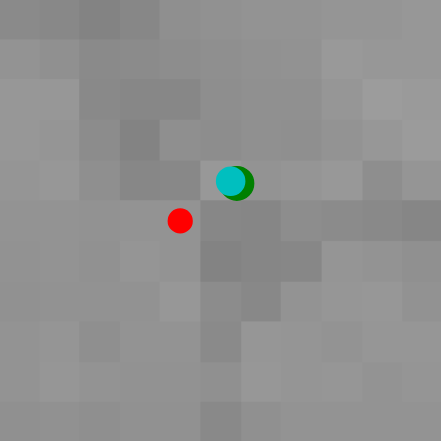}}\\
\subfloat{\includegraphics[width=\x\textwidth]{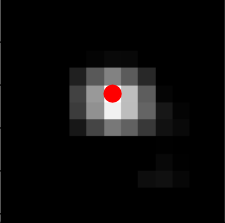}}\hspace{5pt}
\subfloat{\includegraphics[width=\x\textwidth]{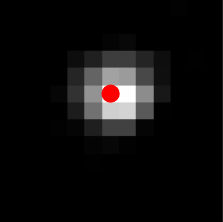}}\hspace{5pt}
\subfloat{\includegraphics[width=\x\textwidth]{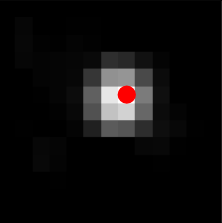}}\hspace{5pt}
\subfloat{\includegraphics[width=\x\textwidth]{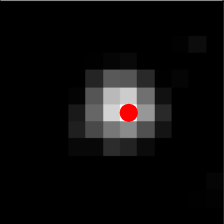}}\hspace{5pt}
\subfloat{\includegraphics[width=\x\textwidth]{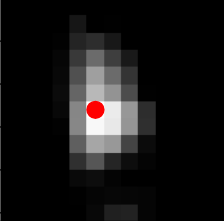}}\hspace{5pt}
\subfloat{\includegraphics[width=\x\textwidth]{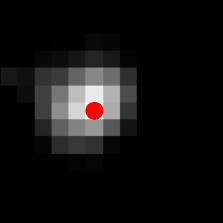}}\\
\subfloat{\includegraphics[width=\x\textwidth]{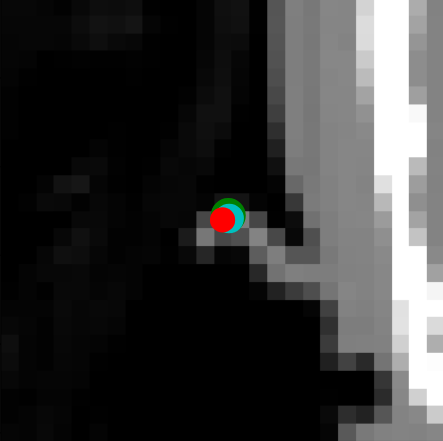}}\hspace{5pt}
\subfloat{\includegraphics[width=\x\textwidth]{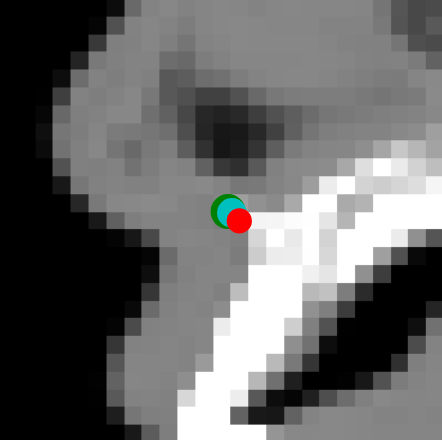}}\hspace{5pt}
\subfloat{\includegraphics[width=\x\textwidth]{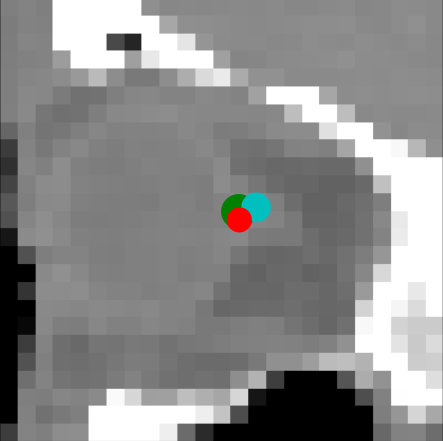}}\hspace{5pt}
\subfloat{\includegraphics[width=\x\textwidth]{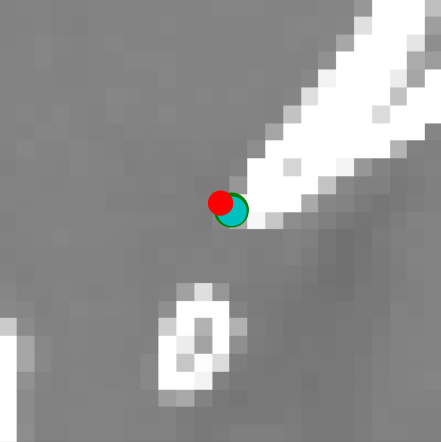}}\hspace{5pt}
\subfloat{\includegraphics[width=\x\textwidth]{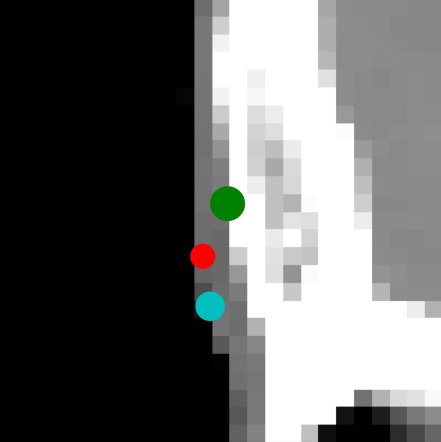}}\hspace{5pt}
\subfloat{\includegraphics[width=\x\textwidth]{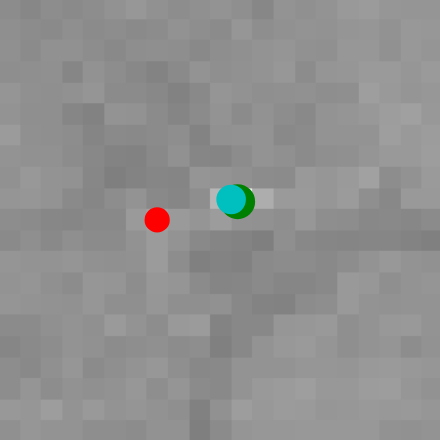}}\\
\addtocounter{subfigure}{-24}
\subfloat[]{\includegraphics[width=\x\textwidth]{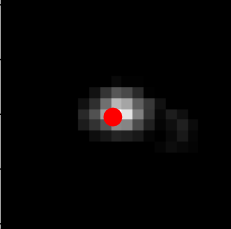}}\hspace{5pt}
\subfloat[]{\includegraphics[width=\x\textwidth]{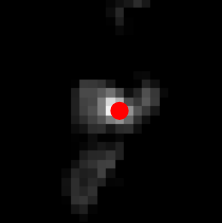}}\hspace{5pt}
\subfloat[]{\includegraphics[width=\x\textwidth]{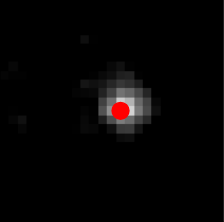}}\hspace{5pt}
\subfloat[]{\includegraphics[width=\x\textwidth]{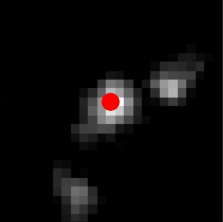}}\hspace{5pt}
\subfloat[]{\includegraphics[width=\x\textwidth]{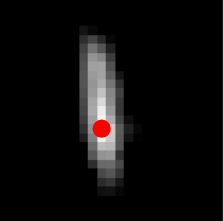}}\hspace{5pt}
\subfloat[]{\includegraphics[width=\x\textwidth]{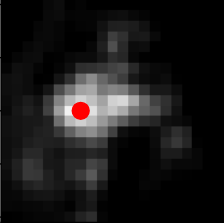}}\\
\vspace{0.2cm}
\caption{A few landmark examples: a) Top of R. ear b) Acanthion c) L. optic nerve d) Opisthion e) Glabella f) Pineal gland. The top row shows a comparison of landmark localisations at full resolution, with green and blue denoting observers A and B and red denoting the FCN detected landmark. The next 4 rows show the detected landmark for Pass 0 and Pass 1, at the 4mm and 2mm algorithmic operating resolutions respectively, along with MIPs of the FCN heatmaps (black = low certainty and white = high certainty). The direct atlas correction is also used in each pass. Slices are taken at the position of observer A, in the sagittal plane for all but the ``Top of R. ear'' where a coronal slice is taken.}
\label{fig:HeadHeatmapResultImages}
\end{figure}

\section{Discussion}

The challenges in this work revolved primarily around how to design a system which could detect multiple structures efficiently in a 3D volume. Even with a relatively modest set of 22 landmarks (note that this is just a subset of the hundreds of landmarks that a whole-body system might be expected to learn), the volume of the outputs and the volume of the final layers of the network is large because of the number of classes, the fact that information is generated at the resolution of the data, and the fact that we work with 3D data. This is in contrast to segmentation tasks with a few classes of interest, to which a network such as U-Net might naturally lend itself. Given this requirement, we designed a system that could be both trained and deployed by making patchwise predictions.

It turned out that with our atlas-assisted detection system to enable the learning of spatial context, a fairly straightforward network with a relatively small receptive field gave good results. From the perspective of deployment, a goal was to be robust to ``awkward'' scan volumes, which might be unusually rotated or cropped, or containing variation due to anatomical or pathological differences. By choosing a model with small receptive field, landmarks are neither reliant nor impacted by spatial context outside of a relatively small neighbourhood. Detection is surprisingly tractable for many landmarks, even with such a limited field of view. Further, so long as we detect sufficient landmarks accurately to compute an accurate affine transform (a minimum of 4 landmarks are required, preferably well spaced \ie not in a planar arrangement), we can leverage the spatial relationships between landmarks to zone in on landmarks with unusual appearance due to pathology, anatomical or postural variation --- albeit without guarantee of precise localisation where pathology has caused an obvious change of appearance. The system also allows mitigation of the time impact of working at higher resolution, by selective evaluation of only the landmark ROIs.



\section{Conclusion}

Convolutional networks have proven their worth for image recognition tasks in general computer vision tasks \cite{Krizhevsky2012,He2015} and in this work, we have shown their efficacy in a medical imaging application, namely the detection of landmarks in head CT volumes. We have benchmarked against a decision forest method (decision forests being the previous gold standard algorithm for this task), for which we have a mature implementation and shown that, given the same system and setup, a neural network significantly outperforms a decision forest, with and without additional feature engineering (\ie HOG features). Further, we have demonstrated that we are able to attain similar agreement to human observers as that between the human observers, showing accuracy that is approximately equal to a \emph{single} human observer.

By exploiting inter-landmark spatial relationships, we are able to use small CNN models with a small receptive field size, and to apply selectively at high resolution. In fact in this paper, we did not show a significant improvement over the simpler system with direct leveraging of an atlas transform alone (our ``atlas correction'' step), and this may be enough to correct outliers and achieve good performance, at least for this problem of landmark detection in head scans. Thus, we have trained a system which is nicely scalable --- to larger scan volumes and to greater numbers of landmarks --- in terms of both GPU memory and run time requirements. The next step is to validate this system on other body parts and other modalities.

\section*{Acknowledgements}

Many thanks to Queen Elizabeth University Hospital, University of Glasgow, who provided many of the medical scans used for this study, including those shown in the images.

\clearpage


\end{document}